%% file: root.tex
\documentclass[letterpaper, 12 pt, journal, twoside, onecolumn]{IEEEtran}
\title{\LARGE \bf
	Variational Inference with Mixture Model Approximation: Robotic Applications 
}

\author{Emmanuel Pignat, Teguh Lembono and Sylvain Calinon
	\thanks{The authors are with the Idiap Research Institute,
		CH-1920 Martigny, Switzerland (e-mail: name.surname@idiap.ch)}%
	\thanks{This work was supported by the MEMMO project (Memory of Motion, http://www.memmo-project.eu/), funded by the European Commission's Horizon 2020 Programme (H2020/2018-20) under grant agreement 780684, and by the Swiss National Science Foundation (CHIST-ERA project I-DRESS, https://i-dress-project.eu/, 20CH21-16085).}
}

\usepackage{graphicx,amsmath,amssymb,array,fancyhdr,color,sidecap,paralist,eurosym,nicefrac,lastpage}
\usepackage[colorlinks=false, pdfborder={0 0 0}]{hyperref}
\usepackage{tikz}
\usepackage{bm}
\usepackage{booktabs}
\usepackage[percent]{overpic}
\usepackage{hyphenat}
\usepackage{color}

\newcommand{\up}{\tilde{p}}
\newcommand{\pa}{\bm{y}}
\newcommand{\Dkl}[2]{D_{\mathrm{KL}}(#1||#2)}

\newcommand{\xd}{\hat{\bm{p}}}

\newcommand{\cost}{\mathrm{c}}

\newcommand{\pmvn}{p_{\mathrm{MVN}}}

\newcommand{\pbmf}{p_{\mathrm{BMF}}}
\newcommand{\trans}[2]{\mathcal{T}_{#2}^{#1}}

\newcommand{\trsp}{{\scriptscriptstyle\top}}

\begin{document}
\maketitle
\bibliographystyle{IEEEtran}

\input{content_corl}

\clearpage




\bibliography{root}  
\newpage
\appendix
\input{appendix_corl}
\end{document}

%% file: content_corl.tex


\begin{abstract}
	We propose a method to approximate the distribution of robot configurations satisfying multiple objectives. Our approach uses variational inference, a popular method in Bayesian computation, which has several advantages over sampling-based techniques. To be able to represent the complex and multimodal distribution of configurations, we propose to use a mixture model as approximate distribution, an approach that has gained popularity recently.
	In this work, we show the interesting properties of this approach and how it can be applied to a wide range of problems in robotics. 
\end{abstract}

\section{Introduction}
%
%
%
%
Many robotic approaches rely on the capability to generate in a quick and efficient way robot configurations fulfilling multiple objectives.
These configurations are typically computed using iterative methods, which can leverage good and diversified initial guesses to improve convergence speed and to find global optimum.

Our work addresses the problem of representing the distribution $p(\bm{x})$ of robot configurations that satisfy multiple given objectives.\footnote{Robot configurations are often denoted $\bm{q}$, but we will avoid this notation because $\bm{q}$ are also used in variational inference to denote approximate distributions.} 
We first show that our goal of representing the distribution of \textit{good and diversified} configurations is the same as the goal pursued by variational inference (VI), a popular and scalable method for Bayesian computation. VI methods approximate an unnormalized density by a tractable one.

As demonstrated in \cite{arenz2018efficient} for a 10-DoF planar robot, we propose to use a mixture model to represent the complex and often multimodal distribution of solutions.

The contributions of the paper are twofold. First, we propose an alternative to Gaussian components more suited to robot kinematics. Then, we extend the framework to conditional distributions using a mixture of experts (MoE) approximation. 
Conditional distributions provide greater online adaptation capabilities to the robot, as it allows us to represent solutions corresponding to distributions of parametrized objectives.
We discuss how several forms of objectives in robotics can be included in this framework, illustrated by various robot applications.
%
%

\section{Two interpretations of the investigated problem}
We start by presenting two interpretations of the problem, both leading to the optimization of the same objective. 
\subsection{Product of experts}
A product of experts \cite{hinton1999products} is a model multiplying several densities $p_m$ (which are called \textit{experts}) together and renormalizing the density. The renormalization makes sure that it is a proper density and introduces dependencies between the experts. Illustratively, each expert $m$ gives its opinion on a different view or transformation $\trans{}{m} (\bm{x})$ of the data $\bm{x}$. Their opinions are then fused together as the product 
\begin{align}
p(\bm{x}|\bm{\theta}_1,\ldots,\bm{\theta}_M) &= 
\frac{\prod_m p_m(\trans{}{m} (\bm{x})\,| \bm{\theta}_m)}
{\int_z \prod_m p_m(\trans{}{m} (\bm{z})\,| \bm{\theta}_m)}.
\end{align}

For example, recalling that $\bm{x}$ denotes the configuration of a robot (joint angles and floating base 6-DoF transformation for a humanoid), $\trans{}{m} (\bm{x})$ can be the position of a foot computed using forward kinematics and $p_m$ a Gaussian distribution defining where the foot should be. The product of experts becomes the distribution of configuration where the foot follows the distribution $p_m$. By adding multiple experts representing various objectives (see Appendix for various examples), the product becomes the distribution fulfilling the set of objectives. 
We point out that the experts do not need to be properly normalized themselves, as the normalization already occurs at the level of the product.
For compactness, we will later refer to the unnormalized product as 
\begin{align}
\up(\bm{x}) &= 
\prod_m p_m(\trans{}{m} (\bm{x})\,| \bm{\theta}_m),
\end{align}
where we dropped the parameters of the experts $\bm{\theta}_1,\ldots,\bm{\theta}_M$ in the notation. 

In the general case, the normalized product $p(\bm{x})$ has no closed-form expression. 
Gaussian experts with linear transformations are a notable exception, where $p(\bm{x})$ is Gaussian, but is of limited interest in our case, due to the nonlinearity of the transformations we are considering. Approximation methods, as used in Bayesian computation, are required; the posterior distribution is indeed the renormalized product on $\bm{\theta}$ of a likelihood and a prior.

Markov chain Monte Carlo (MCMC) is a class of methods to approximate $p(\bm{x})$ with samples. If MCMC methods can represent arbitrary complex distributions, they suffer from some limitations, particularly constraining for our application. They are known not to scale well to high dimension space. 
Except for some methods \cite{chen2014stochastic},
they require an exact evaluation of $\up(\bm{x})$ while stochastic variational inference only requires a stochastic estimate of its gradient. 
MCMC methods also struggle with multimodal distributions and require particular proposal steps to move from distant modes \cite{sminchisescu2003mode}. 
Designing a good proposal step is also algorithmically restrictive. Furthermore, it is difficult to obtain good acceptance rates in high dimension, especially with very correlated $\up(\bm{x})$. When using sampling-based techniques, it is also difficult to assess if the distribution $\up(\bm{x})$ is well covered. 





\subsubsection{Variational Inference}
Variational inference (VI) \cite{wainwright2008graphical} is another popular class of methods that recasts the approximation problem as an optimization. VI approximates the \textit{target density} $\up(\bm{x})$ with a \textit{tractable density} $q(\bm{x}; \bm{\lambda})$, where $\bm{\lambda}$ are the \textit{variational parameters}. \textit{Tractable density} means that drawing samples from $q(\bm{x}; \bm{\lambda})$ should be easy and $q(\bm{x}; \bm{\lambda})$ should be properly normalized. VI tries to minimize the intractable KL-divergence
\begin{align}
\Dkl{q}{p} &= \int_{\bm{x}} q(\bm{x}; \bm{\lambda}) \log \frac{q(\bm{x}; \bm{\lambda})}{p(\bm{x})} d\bm{x}\\
&= \int_{\bm{x}} q(\bm{x}; \bm{\lambda}) \log \frac{q(\bm{x}; \bm{\lambda})}{\up(\bm{x})} d\bm{x} + \log \mathcal{C},
\end{align}
where $\mathcal{C}$ is the normalizing constant. Instead, it minimizes the negative \textit{evidence lower bound} (ELBO) which can be estimated by sampling as
\begin{align}
\mathcal{L}(\bm{\lambda}) &= \int_{\bm{x}} q(\bm{x}; \bm{\lambda}) \log \frac{q(\bm{x}; \bm{\lambda})}{\up(\bm{x})} d\bm{x}\\
\label{equ:elbo_1}
&= \mathbb{E}_{q}[\log q(\bm{x}; \bm{\lambda}) - \log \up(\bm{x})]\\
&\approx \sum_{n=1}^{N}\big(\log q(\bm{x}^{(n)}; \bm{\lambda})-\log \up(\bm{x}^{(n)}) \big)\\
\notag
&\mathrm{with}\quad \bm{x}^{(n)} \sim q(\,.\,|\bm{\lambda})
\end{align}
The reparametrization trick \cite{salimans2013fixed} \cite{ranganath2014black} 
allows us to compute a noisy estimate of the gradient $\mathcal{L}(\bm{\lambda})$, which is compatible with stochastic gradient optimization like Adam \cite{kingma2014adam}. 
For example, if $q$ is Gaussian, this is done by sampling $\bm{\eta}^{(n)}\sim \mathcal{N}(\bm{0}, \bm{I})$ and applying the continuous transformation $\bm{x}^{(n)} = \bm{\mu} + \bm{L}\bm{\eta}^{(n)}$, where $\bm{\Sigma} = \bm{L}\bm{L}^\trsp$ is the covariance matrix. $\bm{L}$ and $\bm{\mu}$ are the \textit{variational parameters} $\bm{\lambda}$. More complex mappings as normalizing flows can be used \cite{rezende2015variational}.

\paragraph{Zero avoiding properties of minimizing $\Dkl{q}{p}$}
\label{par:zero_avoid}
It is important to note that, due to the objective $\Dkl{q}{p}$, $q$ is said to be zero avoiding.
If $q$ is not expressive enough to approximate $\up$, it would miss some mass of $\up$ rather than giving probability to locations where there is no mass (see Fig.~\ref{fig:banana001} for an illustration). 
In our applications, it means that we are more likely to miss some solutions than retrieve wrong ones.
\subsection{Maximum entropy}
Another interpretation is that the robot configurations should minimize a sum of cost 
\begin{align}
	\cost(\bm{x}) &= \sum_{m} \cost_m(\bm{x}),
\end{align}
representing the different objectives.
Computing $\arg\min_{\bm{x}}\cost(\bm{x})$ is equivalent to \textit{maximum a posteriori} in Bayesian statistics. Retrieving a distribution of configurations can be done by finding the distribution $q$ under which the expectation of cost $\mathbb{E}_{q}[\cost(\bm{x})]$ is minimized with the widest entropy $\mathcal{H}(q)$
\begin{align}
\mathcal{L}(\bm{\lambda})&= \mathbb{E}_{q}[\log q(\bm{x}; \bm{\lambda}) + \cost(\bm{x})]\\
&= \mathbb{E}_{q}[\cost(\bm{x})]- \mathcal{H}(q). \label{equ:exp_entropy}
\end{align}
It actually corresponds to \eqref{equ:elbo_1}, where $\up(\bm{x})$ was replaced with 
\begin{align}
	 \up(\bm{x}) &= \exp(-\cost(\bm{x})). 
	 \label{equ:cost_up}
\end{align}

Intuitively, it also fits with our initial objective to generate \textit{good} ($\mathbb{E}_{q}[\cost(\bm{x})]$) and \textit{diversified} ($\mathcal{H}(q)$) samples. 

\section{Mixture model variational distribution}
For computational efficiency, the \textit{approximate distribution} $q(\bm{x}; \bm{\lambda})$ is often chosen as a factorized distribution, using the mean-field approximation \cite{wainwright2008graphical}. 
Correlated distribution can be approximated by a full-covariance Gaussian distribution \cite{opper2009variational}. 
These approaches fail to capture the multimodality and arbitrary complexity of $\up(\bm{x})$. The idea to use a mixture for greater expressiveness as \textit{approximate distribution} is not new \cite{bishop1998approximating}, but the approach has regained popularity recently, see, e.g., \cite{miller2017variational,guo2016boosting,arenz2018efficient}.

A mixture model is built by summing the probability of $k$ mixture components 
\begin{align}
q(\bm{x}|\bm{\lambda}) = \sum_{k=1}^{K} \pi_k\, q_k(\bm{x}|\bm{\lambda}_k), \quad \sum_{k=1}^{K} \pi_k = 1,
\label{equ:mm}
\end{align}
where $\pi_k$ is the total mass of component $k$. The components $q_k$ can be of any family accepting a continuous and invertible mapping between $\bm{\lambda}$ and the samples.
The discrete sampling of the mixture components according to $\pi_k$ has no such mapping. Instead, the variational objective can be rewritten as
\begin{equation}
\mathcal{L}(\bm{\lambda}) =\mathbb{E}_{q}[\log q(\bm{x}; \bm{\lambda}) - \log \up(\bm{x})]
= \sum_{k=1}^{K} \pi_k \mathbb{E}_{q_k}[\log q(\bm{x}; \bm{\lambda}) - \log \up(\bm{x})],
\end{equation}
meaning that we need to compute and get derivatives of expectations only under each \textit{component distribution} $q_k(\bm{x}|\bm{\lambda}_k)$. 
\paragraph{Mixture components distributions}
Gaussian components with full covariance matrix are a natural choice for robotics, due to the quadratic form of its log-likelihood. It can be exploited in standard robotics approaches like linear quadratic tracking (LQR) or inverse kinematics (IK), more details in Sec~\ref{sec:application}.
For some situations, where $\up(\bm{x})$ is very correlated, we propose to use \textit{``banana-shaped''} distribution \cite{haario1999adaptive}, which is done by applying the following differentiable mapping to Gaussian samples $\bm{\eta}^{(n)}\sim \mathcal{N}(\bm{0}, \bm{I})$,
\begin{align}
	f_{\bm{\kappa}}(\bm{\mu} + \bm{L}\bm{\eta}^{(n)}) =
	f_{\bm{\kappa}}(\bm{z}^{(n)}) =
	 \begin{bmatrix}
	z_0^{(n)} + \sum_{i=1}^{N} \kappa_i \, {z_i^{(n)}}^2\\
	z_1^{(n)}\\
	\vdots\\
	z_N^{(n)} 
	\end{bmatrix}.
\end{align}
This mapping can be applied along different parametrized directions. As illustrated in Fig.~\ref{fig:banana001} (b), we get a Gaussian with full covariance, where $\bm{\kappa}$ is a supplementary \textit{variational parameter} encoding curvature.
\subsection{Conditional distribution: Mixture of experts}
Let us suppose the following problem: we want to sample humanoid configurations in static equilibrium within joint limits where the feet are close to given poses. 
As this distribution differs substantially given the poses of the feet, we would like to retrieve this distribution conditioned on the feet locations. That way, no new computation is required according to the current objective, providing very fast online adaptability. More generally, we want to approximate $\up(\bm{x}|\pa)$ where $\pa$ is a \textit{task parameter}, for a distribution of possible $p(\pa)$.

For example, if $\pa_m$ defines a forward kinematics target, expert $p_m$ can be
\begin{align}
	p_m(\bm{x}|\pa_m) \propto \mathcal{N}(\bm{F}(\bm{x}) |\, \pa_m, \sigma\bm{I}).
\end{align}

We propose to use as \textit{approximate distribution} a mixture of experts (ME) 
\begin{align}
q(\bm{x}|\pa;\bm{\lambda}) = \sum_{k=1}^{K} h_{k}(\pa; \bm{\lambda}_k)\, q_k(\bm{x}|\pa;\bm{\lambda}_k), \quad \sum_{k=1}^{K} h_{k}(\pa;\bm{\lambda}_k) = 1,
\label{equ:moe}
\end{align}
where $h_k$ are the gate and $q_k$ are the conditional mixture components as in \eqref{equ:mm}. $q_k$ are normally referred to as \textit{experts}.\footnote{We will avoid this term because of possible confusion with the product of experts.} From \eqref{equ:mm}, we change the mass $\pi_k$ and the \textit{mixture components} $q_k$ to depend on the \textit{task parameters} $\pa$.

In this work, we will use $q_k$ as Gaussian around a linear function, but more complex functions can be considered. Samples from $q_k(\bm{x}|\pa;\bm{\lambda}_k)$ can be generated by first drawing $\bm{\eta}^{(n)}\sim \mathcal{N}(\bm{0}, \bm{I})$ and then applying the mapping $\bm{x}^{(n)} = \bm{W}_k\bm{y} + \bm{c}_k + \bm{L}_k\bm{\eta}^{(n)}$. The \textit{variational parameters} $\bm{\lambda}_k$ are $\bm{W}_k$, $\bm{c}_k$ and $\bm{L}_k$.

Given $\pa$, the objective \eqref{equ:elbo_1} becomes
\begin{align}
\mathcal{L}(\bm{\lambda}, \pa) &= \mathbb{E}_{q}[\log q(\bm{x}|\pa; \bm{\lambda}) - \log \up(\bm{x}|\pa)],
\end{align}
which we would like to minimize under the distribution of possible \textit{task parameters} $p(\pa)$, namely
\begin{align}
\int_{\pa} p(\pa)\mathcal{L}(\bm{\lambda}, \pa) d\pa &= \mathbb{E}_{p(\pa)}\Big[\mathbb{E}_{q}[\log q(\bm{x}|\pa;\bm{\lambda}) - \log \up(\bm{x}|\pa)]\Big]\nonumber\\
&= \mathbb{E}_{p(\pa)}\Big[\sum_{k=1}^{K} h_k(\pa;\bm{\lambda}_k) \mathbb{E}_{q_k(\cdot | \pa;\bm{\lambda}_k)}[\log q(\bm{x}|\pa;\bm{\lambda}) - \log \up(\bm{x}|\pa)]\Big].
\end{align}
This objective can also be minimized by stochastic gradient optimization \cite{salimans2013fixed,ranganath2014black}, by sampling first $\pa^{(n)}\sim p(\pa)$ and then $\bm{x}^{(l, n)}\sim q(\bm{x}|\pa^{(n)}; \bm{\lambda})$ for each $\pa^{(n)}$.


\section{Applications}
\label{sec:application}
\subsection{Learning objectives}
It might seem intuitive to choose the transformations needed to represent a desired joint distribution for a given task, but choosing experts parameters $\bm{\theta_1}, \ldots, \bm{\theta_M}$ can be more complex (if not infeasible).
These parameters can be learned by providing samples of possible desired configurations, for example through kinesthetic teaching.
From a given dataset of robot configurations $\bm{X}$, maximum likelihood (or maximum a posteriori) of the intractable distribution $p(\bm{x}|\bm{\theta}_1,\ldots,\bm{\theta}_M)$ should be computed. 
It can be done using gradient descent \cite{hinton1999products} with
\begin{align}
	\frac{\partial \log p(\bm{x}|\bm{\theta}_1,\ldots,\bm{\theta}_M)}{\partial \bm{\theta_m}} &=
	\frac{\partial\log p_m(\bm{x}|\bm{\theta}_m)}{\partial \bm{\theta_m}} -
	\frac{\partial \log \mathcal{C}(\bm{\theta}_1,\ldots,\bm{\theta}_M)}{\partial \bm{\theta}_m}\nonumber\\
	&= \frac{\partial\log p_m(\bm{x}|\bm{\theta}_m)}{\partial \bm{\theta_m}}\, -
	\int_c p(\bm{c}|\bm{\theta}_1,\ldots,\bm{\theta}_M) \, \frac{\partial\log p_m(\bm{c}|\bm{\theta}_m)}{\partial \bm{\theta_m}} \, d\bm{c}.
	\label{equ:poe_gradient}
\end{align}
Computing a step of gradient requires us to compute an expectation under the intractable distribution $p(\bm{x}|\bm{\theta}_1,\ldots,\bm{\theta}_M)$, making the process computationally expensive. In \cite{hinton2002training}, it is proposed to use a few sampling steps initialized at the data distribution $\bm{X}$. Unfortunately this approach fails when $p(\bm{x}|\bm{\theta}_1,\ldots,\bm{\theta}_M)$ has multiple modes. The few sampling steps never move between modes, resulting in the incapacity of estimating their relative mass. \textit{Wormholes} have been proposed as a solution \cite{welling2004wormholes}, but are algorithmically restrictive. 

Instead, we propose to use VI with the proposed mixture distribution to approximate $p(\bm{x}|\bm{\theta}_1,\ldots,\bm{\theta}_M)$. The training process thus alternates between minimizing $\Dkl{q}{p}$ with current $p$ and using current $q$ to compute the gradient \eqref{equ:poe_gradient}. $q$ can either be used as an importance sampling distribution (or directly if expressive enough to represent $p$). The mass of the modes is directly encoded as $\pi_k$ which makes it much more convenient than moving samples between distant modes.

This process seems overly complex compared to directly encode $p(\bm{x})$, for example as a mixture or a non-parametric density.
However, it offers several advantages, especially when we have access only to small datasets, and/or when very good generalization capabilities are required, namely:
\begin{itemize}
	\item Simple and compact explanation can be found for complex distributions. For example, to learn directly the distribution shown in Fig.~\ref{fig:banana001}, a lot of samples spanning the whole distribution would be required. Finding a simple explanation under a known transformation, following Occam's razor principle, increase generalization capabilities. Very sharp and precise distributions could be obtained with a very limited number of parameters. 
	\item Domain specific a priori knowledge can be included in the form of particular transformations and experts, which would reduce the need for data. More complex transformations can still be learned, if more data are available and more complex objectives are considered.
	\item The model is more structured and interpretable for a human user. It could facilitate interactive or active learning procedure.
\end{itemize}


\subsection{Planning}
Sampling-based robot motion planning \cite{elbanhawi2014sampling} requires sampling of configurations in obstacle-free space, while possibly satisfying other constraints. One of the difficulty is to generate those samples, especially with precise constraints and in high dimension space, which cannot be done with standard rejection sampling approaches. Uniform sampling and projection into the constraint is proposed in \cite{berenson2011task}, and dedicated approaches for closed chains are proposed in \cite{voss2017atlas+,zhang2013unbiased}.
Variational methods can be interesting when high dimension configuration spaces are involved, as they are known to scale better than sampling.

Another challenge of sampling-based planning is to make sure that the space is well covered, ensuring good connectivity. This is indeed a general problem of MCMC methods when approximating a distribution.
Using a mixture model, it is easier to assess for connectivity as the existence of common mass between components. The overlap between two components $k$ and $l$ can be computed for example with 
\begin{align}
	a_{kl}=\int_{\bm{x}} q_k(\bm{x}; \bm{\lambda}_k)\, q_l(\bm{x}; \bm{\lambda}_l)\,d\bm{x},
	\label{equ:overlap}
\end{align}
which is closed-form in case of Gaussian components. Due to the zero avoiding properties of VI, if the components have some common mass, it is very likely that they are connected.

\subsection{Warm-starting inverse kinematics}
Inverse kinematics problems are typically solved with iterative techniques that are subject to local optimum and benefit from good initialization.
It often arises that the poses we find should satisfy general, task-specific and goal-specific constraints. 
For example, a humanoid can achieve a wide range of tasks with its hands while constantly keeping its feet on the ground and ensuring balance.
To be able to warm-start the optimization of all the constraints together, it would be beneficial to pre-compute the distribution of poses, already satisfying the general constraints.


If Gaussian components are used, a nice property is to be able to compute the intersection of objectives without further optimization.
For example, if we have $q_{A}(\bm{x};\bm{\lambda})$ approximating the distribution of poses satisfying the set of objectives $A$, defined by the intractable distribution $\up_{A}(\bm{x})$ and $q_{B}(\bm{x};\bm{\lambda})$ for the set $B$.
We can directly compute $q_{A\cap B}(\bm{x};\bm{\lambda})$, the distribution of configurations satisfying both sets as a product of mixtures of Gaussians, which is itself a mixture of Gaussians \cite{gales2006product}.



%
%

\begin{figure}
	\centering
	\includegraphics[width=.7\linewidth]{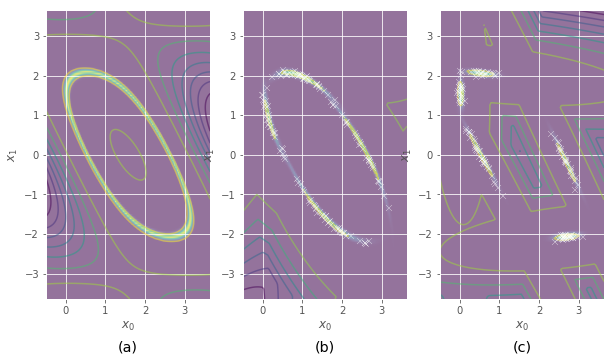}
	\vspace{-10.pt}
	\caption{Precise forward kinematics objectives often create a very correlated distribution of joint angles, which is difficult to sample from or to estimate as a mixture of Gaussians. \textit{(a)} Joint distribution of a 2-DoF planar robot whose end-effector should stay on a line. The colormap represents $\exp(-\cost(\bm{x}))$ and the lines are isolines of $\cost(\bm{x})$. Probability mass is located in yellow (or lighter) areas. \textit{(b)} A mixture of \textit{``banana-shaped''} distributions can substantially reduce the number of mixture components required to represent the distribution, thanks to its capacity to adapt to the local curvature. \textit{(c)} A mixture of Gaussians with the same number of components would not be able to cover efficiently the distribution. Here, for illustrating our point, we employed fewer components than required. The zero-avoiding properties of $\Dkl{q}{p}$ are visible here, as no mass of $q$ is assigned where $\up$ has none.}
	\label{fig:banana001}
\end{figure}

\section{Experiments}
In the experiments, we evaluate the quality of the approximation $q$ with respect to the number of components and the choice of the component distribution. The model with one component is the baseline, which corresponds to the common Gaussian variational approximation (GVA).
The quality of the approximation is evaluated using different similarity and divergence measures between $\up$ and $q$.
These quantities are computed with integrals, which are evaluated by discretizing the configuration space within joint limits.
We consider 3 scenarios: 
\begin{itemize}
    \item a 2-DoF planar robot whose end-effector should stay on a line;
    \item a 7-DoF Panda arm with a given position target for its gripper;
    \item a 28-DoF and floating base Talos humanoid with a given position target of left gripper and feet, where the robot should also ensure static equilibrium.
\end{itemize}
In each one, the robot has more degrees of freedom than required by the task, creating a rich distribution of solutions.

\begin{table}
	\begin{center}
		\caption{Quantitative evaluation of the approximation quality of $q$ for different numbers of components and component distributions. The distribution to approximate $\up$ is shown on Fig.~\ref{fig:banana001} and corresponds to configurations of a 2-DoF planar robot, whose end-effector is on a line. The measures are Bhattacharyya coefficient \cite{bhattacharyya1943measure} and overlapping coefficient \cite{inman1989overlapping} (higher is better, 1 is for perfect reconstruction), Alpha-divergence (or R\'enyi divergence) with $\alpha=1/2$ \cite{renyi1961measures} (lower is better).}
		\begin{tabular}{l|lll|l|lll}
			\toprule 
			 & Bhatt. & OVL & $D_{\alpha=1/2}$ &  & Bhatt. & OVL & $D_{\alpha=1/2}$\\
			\midrule 
			\textit{Banana-shaped} & & &  & \textit{Gaussian} \\ 
			\quad 1 comp. & 0.452& 0.211& 1.096 & GVA - 1 comp. & 0.313& 0.110& 1.374\\
			\quad 5 comp. & 0.793& 0.561& 0.414 &  5 comp. & 0.680& 0.439& 0.640\\ 
			\quad 10 comp. & 0.963& 0.826& 0.073 &  10 comp. & 0.927& 0.738& 0.146\\ 
			\quad 15 comp. & 0.977& 0.854& 0.046 &  15 comp. & 0.948& 0.783& 0.104\\ 
			\quad 20 comp. & 0.995& 0.936& 0.009 &  20 comp. & 0.989& 0.895& 0.022\\  
			\bottomrule 
		\end{tabular}
		\label{table:banana_vs_gaussian}
	\end{center}
\end{table}
Table \ref{table:banana_vs_gaussian} reports the results for the 2-DoF robot and Fig.~\ref{fig:banana001} displays $\up$ and $q$ as colormaps. As expected, one Gaussian is not expressive enough to represent the very correlated distribution created by precise forward kinematic objectives. In a 2-dimensional space, a relatively low number of components is sufficient to approximate $\up$ almost exactly. Models with \textit{``banana-shaped''} component reach quicker the distribution due to their ability to bend.

\begin{table}
	\begin{center}
		\caption{Results for 7-DoF Panda robot (see Table \ref{table:banana_vs_gaussian} for description). The position of the gripper should follow a Gaussian distribution. Sample configurations of these approximations are shown in Fig.~\ref{fig:pandagmmvsgaussian}.}
		\begin{tabular}{l|ll|l|ll}
			\toprule 
			& Bhatt. & $D_{\alpha=1/2}$ &  & Bhatt. & $D_{\alpha=1/2}$\\
			\midrule 
			\textit{``Banana-shaped''} & &  & \textit{Gaussian} \\ 
			\quad 1 comp. & 0.091& 1.81 & GVA - 1 comp. & 0.085& 1.89\\
			\quad 10 comp. & 0.362& 1.27 &  10 comp. & 0.225& 1.55\\ 
			\quad 100 comp. & 0.587 & 0.82 &  100 comp. & 0.505& 0.98\\  
			\bottomrule 
		\end{tabular}
			\vspace{-15.pt}
		\label{table:banana_vs_gaussian_panda}
	\end{center}
\end{table}

\begin{figure}
	\centering
	\includegraphics[width=0.7\linewidth]{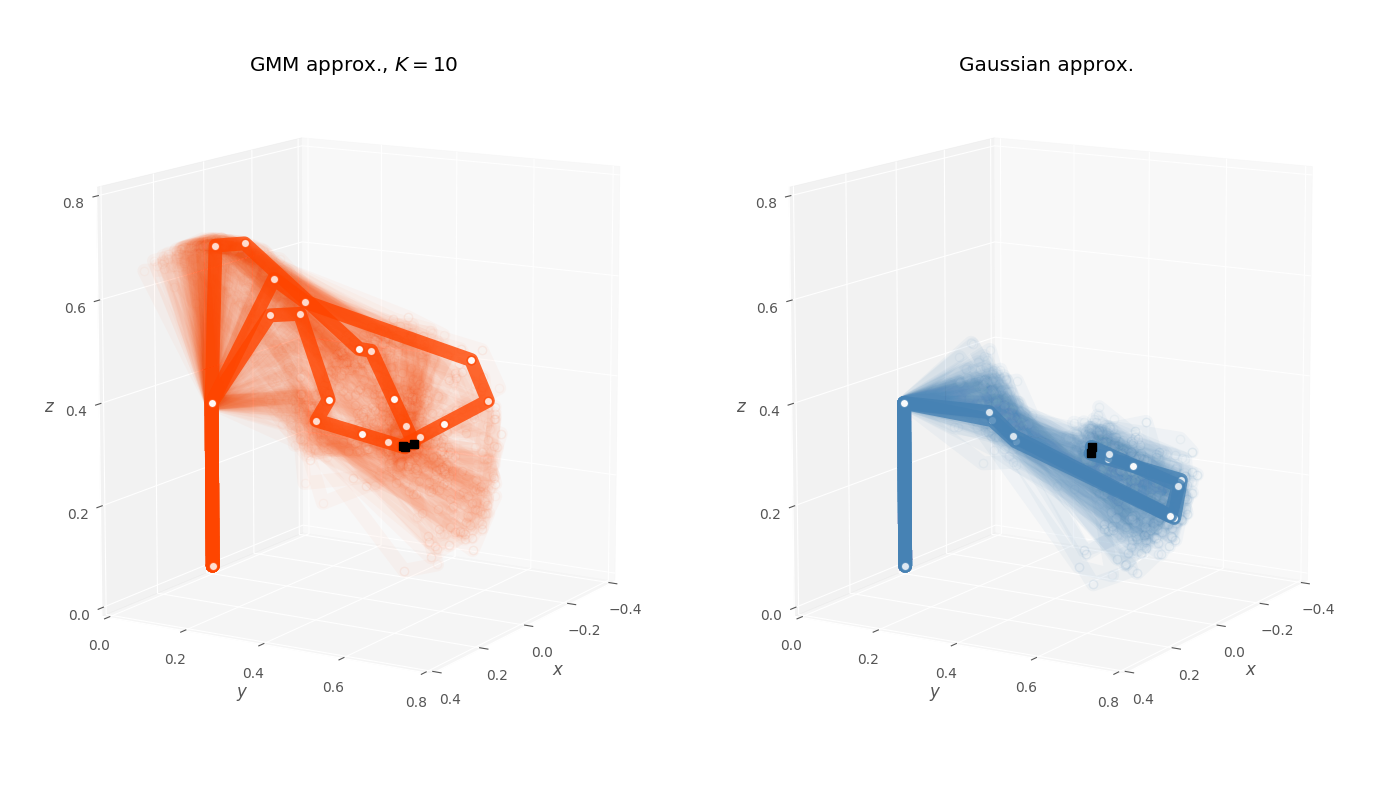}
	\vspace{-20.pt}
	\caption{Comparing approximations $q$ of the distribution of configuration of the 7-DoF Panda robot whose end-effector position follows a Gaussian distribution. \textit{(left)} Gaussian mixture model approximation with $K=10$ components. \textit{(right)} Gaussian approximation. For each approximation, 3 samples drawn from the distribution are shown as well as 200 more samples in transparency. The GMM approximation covers a significantly higher proportion of the distribution and proposes much more diversified solutions.}
	\label{fig:pandagmmvsgaussian}
\end{figure}

Table \ref{table:banana_vs_gaussian_panda} reports the results for the 7-DoF Panda robot. Because of the 7 dimensions, the distribution is significantly richer and more complex to represent. If the results suggest that we are not representing the distribution as well as in the previous scenario, Fig.~\ref{fig:pandagmmvsgaussian} shows that we already have a quite good variety of configurations with 10 Gaussian components. For a better approximation, an iterative training procedure can be used, such as proposed in \cite{miller2017variational,guo2016boosting}. It incrementally builds a richer approximation by increasing the number of states.
Compared to sampling, an interesting feature of the proposed method is that the insensitivity of the task with respect to the last joint of the robot (the wrist) can be directly encoded in the covariance matrices. This greatly reduces the number of units (samples or clusters) to represent the distribution, which can be crucial in high-dimension space.

\begin{figure}
	\centering
	\includegraphics[width=.7\linewidth]{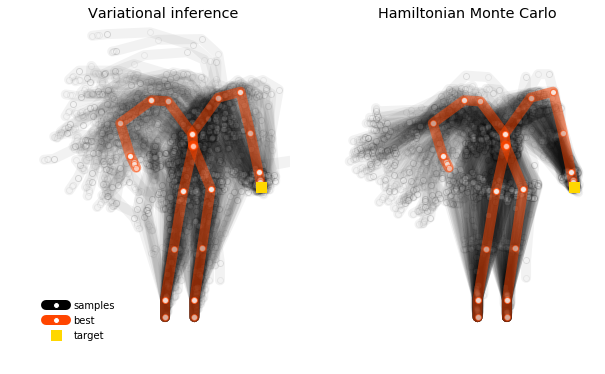}
	\vspace{-16.pt}
	\caption{Samples of configurations where the Talos robot should have its left hand at a given position and both feet at a given pose. A static equilibrium should also be ensured. \textit{(left)} 100 samples from GMM variational distribution with $K=50$ components. \textit{(right)} 100 randomly picked samples of a 2000 steps Hamiltonian Monte Carlo chain. Both learning procedures were prematurely stopped. It illustrates the advantages of the variational inference by its efficiency (in terms of time and memory) to represent irrelevant dimensions.}
	\label{fig:talosvariationalvssampling001}
\end{figure}

For the humanoid robot, running evaluations using discretization is too expensive. We show in Fig.~\ref{fig:talosvariationalvssampling001} samples of configuration from $q$, a Gaussian mixture model with $K=50$ components, using Hamiltonian Monte Carlo (HMC). We allowed for a limited 40 sec.\ of stochastic optimization for variational inference and HMC steps (equivalent to 2000 steps).
For HMC, the kernel was chosen Gaussian with a standard deviation of $0.1$. Fig.~\ref{fig:talosvariationalvssampling001} shows a higher variance of the right arm for variational inference. The related degrees of freedom have little to no influence on the task (only for balancing). Thanks to the encoding of Gaussian covariances and the maximization of entropy in \eqref{equ:exp_entropy}, the method can quickly localize unimportant dimensions. Compared to the sampling approach, it has benefits both in terms of computational and memory efficiency, as well as in generalization capabilities.

\section{Conclusion}
We have proposed the use of variational techniques with mixture models as a general tool to represent distributions of robot configurations satisfying several forms of objectives. We emphasized its advantages over sampling methods. Computational and memory requirements can be significantly reduced in high dimension space, which can be crucial in many applications. 

We proposed several robotic applications of this method, which deserve to be investigated in more details in future work. It is particularly promising for training unnormalized density (e.g., energy-based model or PoE) as it can handle multimodality better than sampling approaches.

%% file: appendix_corl.tex
In this document, several transformations $\trans{m}{}$ and experts models $p_m$ related to common robotic problems are presented as a cookbook.
\subsection{Transformations}
\label{sec:exps_and_costs}
\paragraph{Forward kinematics (FK)} One of the most common transformation used in robotics is forward kinematics, computing poses (position and orientation) of links given the robot configuration $\bm{x}$. Using variational inference with a mixture of Gaussians was already proposed in \cite{arenz2018efficient} to find all the distribution of configurations of a 10 DoF planar robot where the end link is close to a target.

Forward kinematics can also be computed in several task spaces associated with objects of interest \cite{calinon2016tuto}. 
\paragraph{Center of Mass (CoM)}
From the forward kinematics of the center of mass of each link and their mass, it is possible to compute the center of mass (CoM) of the robot. To satisfy static equilibrium, the CoM should be on top of the support polygons. 

\paragraph{Distance} A \textit{relative distance space} is proposed in \cite{yang2015real}. It computes the distances from multiple virtual points on the robot to other objects of interests (targets, obstacles). It can for example be used in environments with obstacles, providing an alternative or complementing forward kinematics.

\paragraph{Steps of Jacobian pseudo-inverse}
Precise kinematics constraints imply a very correlated $\up(\bm{x})$ from which it is difficult to sample and represent as a mixture of Gaussians. In the extreme case of hard kinematics constraints, the solutions are on a low dimensional manifold embedded in configuration space. Dedicated methods address the problem of representing \cite{voss2017atlas+} or sampling this manifold \cite{zhang2013unbiased}.
In \cite{berenson2011task}, a \textit{projection} strategy is proposed. Configurations are sampled randomly and projected back using an iterative process. We propose a similar approach where the projection operator $\mathcal{P}_N$ would be used as transformation $\trans{m}{}$.
Inverse kinematics problems are typically solved iteratively with
\begin{align}
	\mathcal{P}(\bm{x}) &=\bm{x} + J(\bm{x})^\dagger \Big(\xd - \bm{F}(\bm{x})\Big),
	\label{equ:ik_iteration}
\end{align} 
where $\hat{\bm{p}}$ is the target and $J(\bm{x})^\dagger$ is the Moore-Penrose pseudo-inverse of the Jacobian.
This relation is derivable and can be applied recursively with
\begin{align}
	\mathcal{P}_{0}(\bm{x}) &= \mathcal{P}(\bm{x}),\\
	\mathcal{P}_{n+1}(\bm{x}) &= \mathcal{P} \Big(\mathcal{P}_{n-1}(\bm{x})\Big).
\end{align} 

\begin{figure}
	\centering
	\includegraphics[width=.7\linewidth]{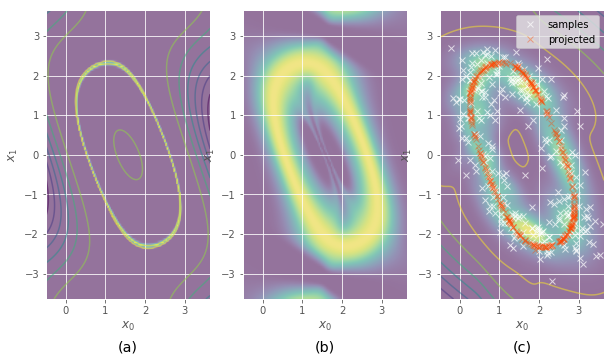}
	\caption{Precise forward kinematics objectives often create a very correlated distribution of joint angles, which is difficult to sample from or to estimate as a mixture of Gaussians. 
	\textit{(a)} Joint distribution of a 2-joint robot whose end-effector should stay on a line. The colormap represents $\exp(-\cost(\bm{x}))$ and the lines are isolines of $\cost(\bm{x})$. Probability mass is located in yellow (or lighter) areas. 
	\textit{(b)} Distribution of points from which two iterations of IK leads to distribution (a). 
	\textit{(c)} Its approximation as a mixture of Gaussians with samples and projected samples with two iterations.}
	\label{fig:ik_001}
\end{figure}

Then, the distribution
\begin{align}
p_m(\bm{x})\propto\mathcal{N}\Big(\mathcal{P}_{N}(\bm{x})|\, \xd, \sigma \bm{I}\Big)
\end{align}
is the distribution of configurations which converge in $N$ steps to $\mathcal{N}(\xd, \sigma \bm{I})$, see Fig.~\ref{fig:ik_001}. Thanks to the very good convergence of the iterative process \eqref{equ:ik_iteration}, $\sigma$ can be set very small.
However, this approach has a similar (but less critical) problem as \cite{berenson2011task}. The \textit{approximate distribution} $q(\bm{x}; \bm{\lambda})$ will be slightly biased toward zones where the forward kinematics is close to linear (constant Jacobian), which are those where more mass converge to the manifold. 

With high DoF robots, it might be computationally expensive to run iteration steps inside the \textit{stochastic gradient optimization} and propagate the gradient. Another approach would be to define heuristically, or learn, $\bm{\Sigma}_h$ such that $\mathcal{N}(\bm{x}|\, \xd, \sigma \bm{I} + \bm{\Sigma}_h)$ is close to 
$\mathcal{N}(\mathcal{P}_{N}(\bm{x})|\, \xd, \sigma \bm{I})$.

\section{Distributions}
We present several distribution that can be used as cost on the transformations.
\paragraph{Multivariate normal distribution (MVN)}
An obvious choice for forward kinematics objective is the Gaussian or multivariate normal distribution (MVN). Its log-likelihood is quadratic, making it compatible with standard inverse kinematics and optimal control techniques,
\begin{equation}
\pmvn(\bm{X} |\bm{\mu}, \bm{\Sigma})\propto 
\exp\Big(-\frac{1}{2}(\bm{x}-\bm{\mu})^{\trsp}
\bm{\Sigma}^{-1}(\bm{x}-\bm{\mu})\Big),
\end{equation}
where $\bm{\mu}$ is the location parameter and $\bm{\Sigma}$ is the covariance matrix.

\paragraph{Matrix Bingham-von Mises-Fisher distribution (BMF)}
To cope with orientation, for example represented as rotation matrix, Matrix Bingham-von Mises-Fisher distribution (BMF) \cite{khatri1977mises} can be used. Its normalizing constant is intractable and requires approximation \cite{kume2013saddlepoint}, which is not a problem in our case, as we integrate over robot configurations.
Its density
\begin{equation}
\pbmf(\bm{X} |\bm{A}, \bm{B}, \bm{C} )\propto 
\exp\Big(\operatorname {tr} (\bm{C} ^{\trsp}\bm{X} + 
\bm{B}\,{\bm{X}}^{\trsp} \bm{A}\, \bm{X})\Big),
\end{equation}
has a linear and a quadratic term and can be written as an MVN with a particular structure of the covariance matrix. Correlations between rotations can be represented with this distribution. Rewritten as an MVN, it is possible to create a joint distribution of position and rotation matrices that also encodes correlations between them.
\paragraph{Cumulative distribution function (CDF)}
Inequality constraints, such as static equilibrium, obstacles or joint limits can be treated using cumulative distribution function
\begin{align}
	p(x \leq b),\quad \mathrm{with}\quad x \sim \mathcal{N}\Big(\trans{}{}(\bm{x}), \sigma^2\,\Big),
	\label{equ:cdf}
\end{align}
where $\trans{}{}(\bm{x})$ is a scalar. For example for half-plane constraints, $\trans{}{}(\bm{x})$ could be $\bm{w}^\trsp\bm{x}$, or for joint limits on first joint $\trans{}{}(\bm{x})=\bm{x}_0$.

The use of the CDF makes the objectives continuous and allows us to consider a safety margin determined by $\sigma$.

Obstacles constraints might be impossible to compute exactly and require collision checking techniques \cite{elbanhawi2014sampling}. Due to the stochastic optimization, our approach is compatible with stochastic approximation of the collision related cost, which might speed up computation substantially.

\paragraph{Uni-Gauss distribution}
To represent hierarchy between multiple tasks in our framework, we propose to use uni-Gauss experts \cite{hinton1999products}. It combines the distribution defining a non-primary objective $p_m$ with a uniform distribution
\begin{align}
p_{\mathrm{UG},m}(\bm{x}) = \pi_m p_m(\bm{x}) + \frac{1-\pi_m}{c},
\label{equ:unigauss}
\end{align}
which means that each objective has a probability $p_m$ to be fulfilled. It can also be interpreted as a cost of $\log(1-\pi_m)$ penalizing the neglect of task $m$.

Classical prioritized approaches \cite{nakamura1987task} exploit spatial or temporal redundancies of the robot to achieve multiple tasks simultaneously. They use a null-space projection matrix, such that commands required to solve a secondary task do not influence the primary task.
As our tasks are defined as distributions, we do not necessarily need redundancies. As each non-primary tasks has a probability to be abandoned, approximating the PoE would evaluate if there is sufficient mass at the intersection of the objective.

There are two possible ways of estimating $\up(\bm{x})$ in case of Uni-Gauss experts. If the number of tasks is small, we can introduce, for each task $m$, a binary random variable indicating if the task is fulfilled or not. For each combination of these variable, we can then compute $\up(\bm{x})$. The ELBO can be used to estimate the relative mass of each of these combinations, as done in model selection. For example, if the tasks are not compatible, their product would have a very small mass, as compared to the primary task. In the case of numerous objectives, this approach becomes computationally expensive because of the growing number of combinations. We can instead marginalize these variables and we fall back on \eqref{equ:unigauss}. For practical reasons of flat gradients, the uniform distribution can be implemented as a distribution of the same family as $p_m$ with a higher variance, that changes across optimization.

\begin{figure}
	\centering
	\includegraphics[width=1.\linewidth]{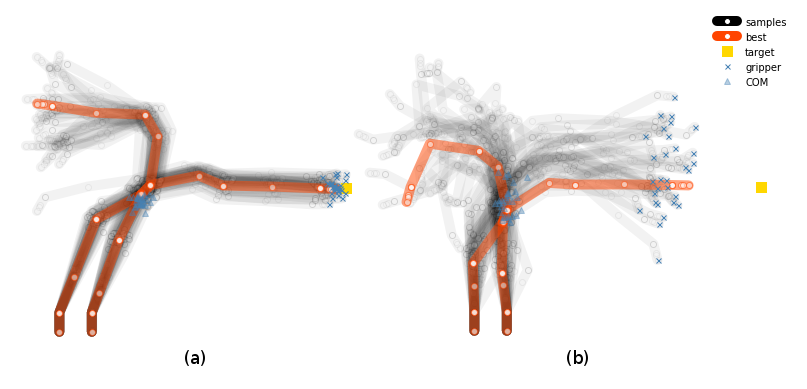}
	\caption{Distribution of poses satisfying static equilibrium, joint limits, feet and left gripper IK objectives. Feet are reprojected to their exact locations using three Newton's method steps. \textit{(a)} No hierarchy between the different objectives. \textit{(b)} Left gripper is set as a secondary objective using Uni-Gauss expert. Given that the secondary objective has some probability not to be fulfilled, there is more mass in the intersection of the other constraints than in the intersection of all.}
	\label{fig:COM_secondary}
\end{figure}

Fig.~\ref{fig:COM_secondary} (a) shows an example where forward kinematics objectives are given for the feet and a hand while static balance has to be ensured. 

\subsection{Nullspace}
The proposed approach is also compatible with classical null-space formulations \cite{nakamura1987task}. We can exploit the fact that stochastic variational inference does not need to evaluate the unnormalized density $\up$ but only its gradient. This characteristic is shared with only a very few Monte Carlo methods such as \cite{chen2014stochastic}. It allows us to filter the gradient of the different objectives (expert and transformation). The derivative of the expert log-probability with respect to the configuration $\bm{x}$ can be written as 
\begin{align}
\frac{\partial \log p(\trans{}{m}(\bm{x})|\bm{\theta}_m)}{\partial \bm{x}} &=
\frac{\partial \log p(\bm{y}|\bm{\theta}_m)}{\partial \bm{y}}  
\frac{\partial \trans{}{m}(\bm{x})}{\partial \bm{x}}\\
&=\frac{\partial \log p(\bm{y}|\bm{\theta}_m)}{\partial \bm{y}}  
\bm{J}_m(\bm{x}),
\end{align}
where we make the Jacobian $\bm{J}_m(\bm{x})$ of the transformation appear. When multiple tasks should be prioritized, a null-space filter $\bm{N}_1(\bm{x})$ can be added such that the gradient of the secondary objective only acts perpendicularly to the main objective, namely
\begin{align}
\frac{\partial \log p(\trans{}{m}(\bm{x})|\bm{\theta}_1, \bm{\theta}_2)}{\partial \bm{x}}
=
\frac{\partial \log p(\bm{y}|\bm{\theta}_1)}{\partial \bm{y}}  
\bm{J}_1(\bm{x}) + \frac{\partial \log p(\bm{y}|\bm{\theta}_2)}{\partial \bm{y}} \bm{J}_2(\bm{x}) \bm{N}_1(\bm{x}),\\
\quad \mathrm{where}\quad \bm{N}_1(\bm{x}) = \bm{I} - \bm{J}_1(\bm{x})^\dagger \bm{J}_1(\bm{x}).
\end{align}
Note that when using automatic differentiation library, gradients can be easily redefined with this filter.

\begin{figure}
	\centering
	\includegraphics[width=1.\linewidth]{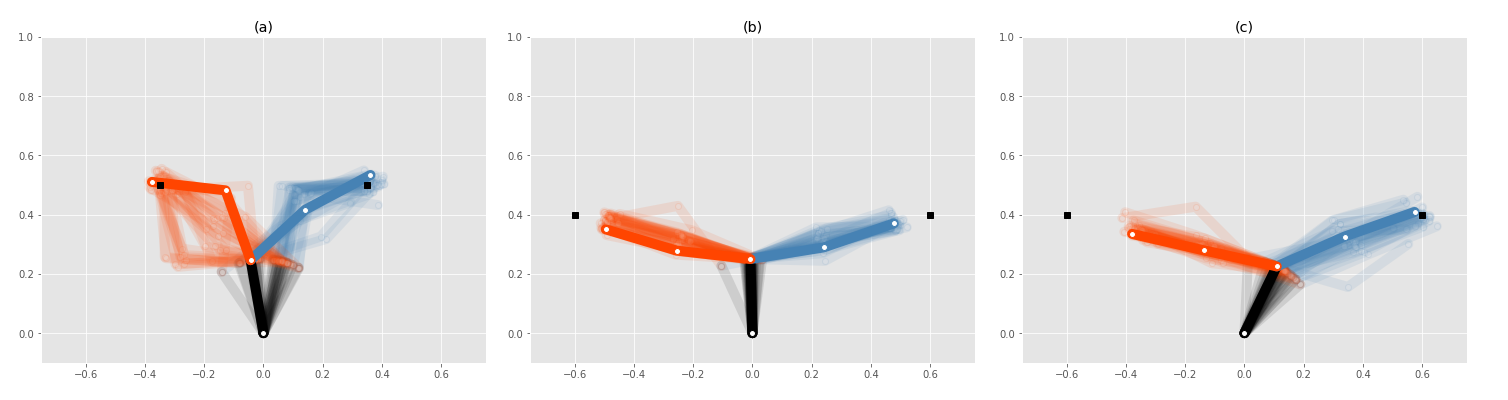}
	\caption{5 DoF bimanual planar robot with two forward kinematics objectives. \textit{(a)} The two tasks are compatible and the distribution of solution is approximated. \textit{(b)} No null-space, the two tasks are of the same importance. \textit{(c)} The gradient of the objective of the orange arm is projected onto the null-space of the Jacobian of the forward kinematics of the blue arm. This is made possible because stochastic variational inference only requires to evaluate the gradient of the unnormalized density. }
	\label{fig:nullspace}
\end{figure}

Fig.~\ref{fig:nullspace} shows a 5 DoF bimanual planar robot with two forward kinematics objectives. When the tasks are compatible, the filtering has no effect. The gradient of the objective of the orange arm can be projected onto the null-space of the Jacobian of the forward kinematics of the blue arm, resulting in a prioritization.


